\pdfoutput=1
\documentclass[11pt]{article}

\usepackage{EACL2023}
\usepackage{perpage}

\usepackage{times}
\usepackage{latexsym}
\usepackage{graphicx}

\usepackage[T1]{fontenc}
\usepackage[utf8]{inputenc}

\usepackage{microtype}

\usepackage{inconsolata}

\usepackage{algpseudocode}
\usepackage{bm}
\usepackage{booktabs}
\usepackage{todonotes}
\title{Multilingual Gradient Word-Order Typology from Universal Dependencies}

\author{Emi Baylor\textsuperscript{*} \\
  McGill University \\
  Mila Quebec AI Institute \\
  \texttt{\small emily.baylor@mail.mcgill.ca} \\\And
  Esther Ploeger\textsuperscript{*} \\
  Dept.~of Computer Science \\
  Aalborg University \\
  \texttt{\small espl@cs.aau.dk} \\\And
  Johannes Bjerva \\
  Dept.~of Computer Science \\
  Aalborg University \\
  \texttt{\small jbjerva@cs.aau.dk} \\}

\begin{document}
\maketitle
\let\thefootnote\relax\footnotetext{* These authors contributed equally to this work.}

\begin{abstract}
While information from the field of linguistic typology has the potential to improve performance on NLP tasks, reliable typological data is a prerequisite.
Existing typological databases, including WALS and Grambank, suffer from inconsistencies primarily caused by their categorical format. 
Furthermore, typological categorisations by definition differ significantly from the continuous nature of phenomena, as found in natural language corpora. In this paper, we introduce a new seed dataset made up of continuous-valued data, rather than categorical data, that can better reflect the variability of language. While this initial dataset focuses on word-order typology, we also present the methodology used to create the dataset, which can be easily adapted to generate data for a broader set of features and languages.
\end{abstract}

\section{Introduction}
\label{sec:introduction}

Data from the field of linguistic typology has the potential to be useful in training NLP models \citep{bender2016-paper2,ponti2019modeling}. However, the main existing typological databases, WALS (World Atlas of Language Structures) \citep{wals} and Grambank \citep{grambank_2023}, contain inconsistent and contradictory information \citep{baylor_underReview}. These issues stem, in large part, from the categorical format of the data, which is over-simplistic and therefore cannot capture the nuance and variability that exist in natural language.

For example, one of the features describes the ordering of adjectives and the noun they modify. The categories in these datasets are Noun-Adjective, Adjective-Noun, or Variable. Limiting the options to these three categories removes any information differentiating a language that employs Noun-Adjective ordering 10\% of the time from one that does so 90\% of the time. In addition, the threshold between the Noun-Adjective and Adjective-Noun categories and the Variable category is often not clear, which can lead to inconsistencies in the data. As an example, the same 90\% Noun-Adjective language might be classified as Variable in one database, but might be seen as consistently Noun-Adjective enough to be classified in the Noun-Adjective category in another database.

\begin{figure}[t]
    \centering
    \includegraphics[width=0.5\textwidth]{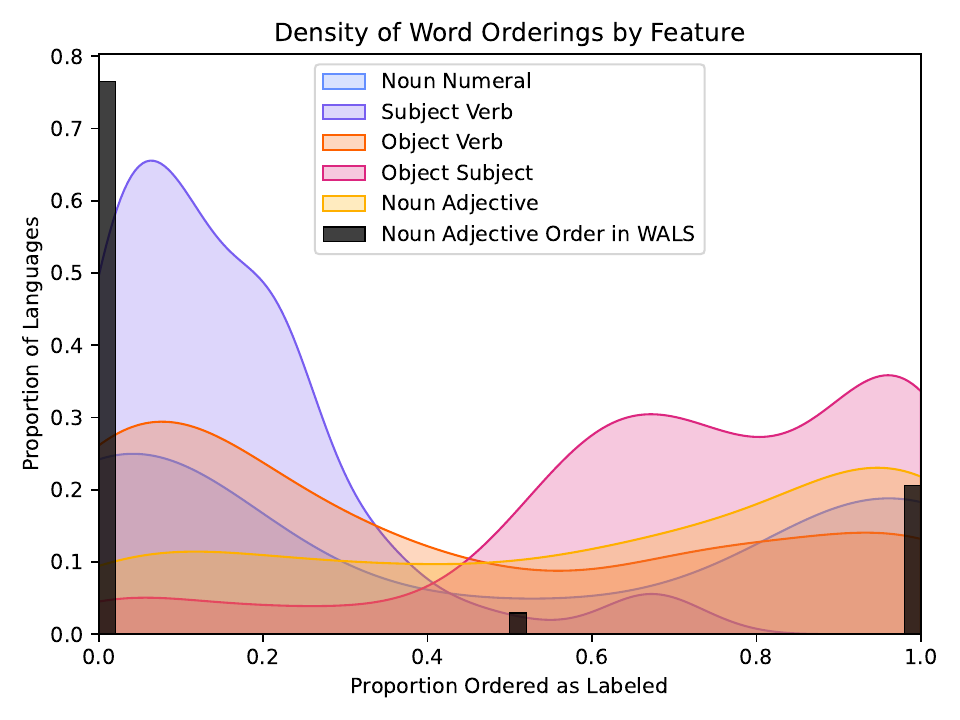}
    \caption{Proportion of languages with proportion of relevant words ordered as labeled, by feature. The black represents WALS Noun Adjective categories, with the far left being the Adjective Noun languages, the far right being the Noun Adjective languages, and the center being the variable languages. All other distributions come from our dataset.}
    \label{fig:featureDensity}
\end{figure}

% below is table for 'dataset creation'
\begin{table*}[!ht]
\centering
\resizebox{0.98\textwidth}{!}{
\begin{tabular}{lccl}
    \toprule
        \bf{French phrase} & \bf{\textcolor{Purple}{Noun}-\textcolor{Red}{Adjective} Count} & \bf{\textcolor{Red}{Adjective}-\textcolor{Purple}{Noun} Count} & \bf{English Translation}  \\
    \midrule
        Mon \textcolor{Red}{\textit{cher}} \textcolor{Purple}{\underline{ami}} & 0 & 1 & \textit{"My dear friend"} \\
        Mon \textcolor{Purple}{\underline{appartement}} \textcolor{Red}{\textit{ancien}} & 1  & 0  & \textit{"My ancient apartment"} \\
    \bottomrule
\end{tabular}}
\caption{An example of counting Noun-Adjective and Adjective-Noun instances in the dataset creation process, with English translations for ease. French nouns are underlined and in \textcolor{Purple}{\underline{purple}}, and French adjectives are italicized and in \textcolor{red}{\textit{red}}.}
\label{tab:nounAdjCount}
\end{table*}

In this paper, we apply recommendations presented in \citet{levshina_2023} and extend their analysis by introducing a new continuous-valued typological dataset that removes the need to oversimplify data into categories. In particular, we focus on word-level typology, and present a method for extracting gradient typology that utilizes the part of speech annotations available in the Universal Dependencies (UD) treebank corpus \citep{nivre-etal-2020-universal}. We then propose a novel regression-based typology task. 

This new dataset and the methods used to create it are beneficial not only to NLP, but also potentially to the field of linguistic typology itself.
Similar to previous works that include automatically recognizing or utilizing typological information \citep{asgari-schutze-2017-past, saha-roy-etal-2014-automatic, nikolaev-pado-2022-word}, we introduce a new data collection technique that can methodically extract typological information from existing annotated text-based datasets.

\section{Background}
\label{sec:background}

\subsection{Linguistic Typology}
\label{sec:typology}
Linguistic typology is the study of the world's languages through the comparison of specific features of language, across a variety of languages \citep{ponti2019modeling, comrie1988linguistic}. These features can focus on any aspect of language, including phonology \citep{hyman2008universals,lindblom1988phonetic}, syntax \citep{greenberg1966,comrie:1989}, morphology \citep{comrie:1989}, and phonetics \citep{lindblom1988phonetic}.

For example, a typologist might look to contrast the number of distinct vowels that a diverse group of languages employ \citep{wals-2}. 
Or they would compare how different languages tend to order verbs and their subjects: do verbs generally occur before or after their subjects in a sentence? \citep{wals-82}.
Compared to other areas of linguistics, word order data can be relatively easy to collect, meaning that word order features tend to have data across a large number of the world's languages. 
Additionally, within NLP, word-order is the most commonly studied typological feature when it comes to cross-lingual transfer \citep{philippy-etal-2023-towards}.
Typological diversity is furthermore used in NLP as an argument for language sampling, albeit without any consensus for the underlying meaning of the term \citep{ploeger-etal-typdiv-2024}.

\subsection{Existing Typological Resources}
\label{sec:typResources}

The current two most popular typological databases, WALS \citep{wals} and Grambank \citep{grambank_2023}, offer coverage of over 2,000 languages each. 
While the overall scope of the databases is great, their reliance on categorical representations of linguistic features means they frequently oversimplify data to the point of creating inconsistencies and errors, both within the databases, and with each other. 
Although this categorical distinction is a conscious design choice, we argue that a data driven and gradient solution can provide benefits both for typology and for NLP.

One solution to this problem of discrete categorical representations, proposed by \citet{levshina_2023}, is to instead replace them with gradient representations. These continuous gradient representations are better able to capture nuanced linguistic information.

\section{Continuous-Valued Seed Dataset}
\label{sec:dataset}

We introduce a seed dataset based on the idea of continuous representations of linguistic features \citep{levshina_2023}. This dataset is currently small, with coverage of fewer than 100 languages across a limited number of features. However, the process used to create it, described in section \ref{sec:datasetCreation}, can be easily adapted for broader feature coverage, as well as broader language coverage.

\subsection{Dataset Creation}
\label{sec:datasetCreation}
To best describe the creation of this dataset, we will walk through the data collection process for a single linguistic feature: the relative orderings of adjectives and the nouns they modify. In WALS (feature 87A) and Grambank (feature GB025), the ordering of nouns and adjectives are represented categorically, with languages generally split into three categories: Adjective-Noun, Noun-Adjective, or No dominant order. Instead of trying to fit a given language into one of these discrete categories, we extract the proportions of Adjective-Noun and Noun-Adjective instances in that language's Universal Dependencies (UD) treebank \citep{nivre-etal-2020-universal}. 

To do this, we iterate through all of the sentences in the given dataset, counting the number of times adjectives occur before the noun they modify, as well as the number of times they occur after the noun they modify. Two examples can be seen in Table \ref{tab:nounAdjCount}, where the phrase \textit{Mon cher ami} adds one to the Adjective-Noun count, and the phrase \textit{Mon appartement ancien} adds one to the Noun-Adjective count. We then use those counts to calculate the proportion of Adjective-Noun vs. Noun-Adjective instances that occur in the dataset.

We repeat this process for every dataset in UD that includes the necessary Noun and Adjective part of speech annotations. This algorithm is described in pseudocode in Figure \ref{alg:datasetCreation}. Because some languages have multiple datasets in UD, these languages have multiple Adjective-Noun and Noun-Adjective proportion datapoints. In the case of our seed dataset, we were able to extract information from 132 different UD datasets, within which there are 91 unique languages.

For this seed dataset, we extract data for five features:
\begin{enumerate}
    \item Ordering of adjectives and their nouns
    \item Ordering of numerals and their nouns
    \item Ordering of subjects and verbs
    \item Ordering of objects and verbs
    \item Ordering of objects and subjects
\end{enumerate}

Each feature required manual adjustments of the dataset creation code in order to extract the necessary part of speech information from the annotated UD data. These changes are small overall, generally requiring only an adjustment of the UD tags being matched. The tags we used can be found in Table \ref{tab:ud-tags} of Appendix \ref{sec:appendix}.

\begin{figure}
\colorbox[RGB]{239,240,241}{
   \begin{minipage}{.9\linewidth}
    \begin{algorithmic}
    
    \ForAll{$d \in \mbox{UD Datasets}$}
        \State $na \gets 0$
        \Comment{$na$ is the Noun-Adj count}
        \State $an \gets 0$
        \Comment{$an$ is the Adj-Noun count}
        \ForAll{$\mbox{sentence } s \in d$}
            \State $na \gets na+$ \textbf{count }Noun-Adj in $s$
            \State $an \gets an+$ \textbf{count }Adj-Noun in $s$
        \EndFor
    \State $na\_proportion \gets \frac{na}{na+an}$
    \EndFor
    \end{algorithmic}
     \end{minipage}     }
          \caption{Pseudocode depicting our process of collecting data for one linguistic feature.}

    \label{alg:datasetCreation}
\end{figure}

\subsection{Value Distributions}
\label{sec:valueDistributions}

As Figure \ref{fig:featureDensity} demonstrates, each feature's data creates a different distribution across the range of possible proportions. Using these raw proportions allows us to observe linguistic differences between languages that would previously be collapsed into the same category. 
This is made especially clear by the visualization of WALS data (black) in Figure \ref{fig:featureDensity}, which is a much more limited distribution than its Noun Adjective counterpart in yellow.

\begin{table*}[!ht]
    \centering
    \resizebox{0.94\textwidth}{!}{
    \begin{tabular}{l|cc|cc}
    \toprule
         & \textbf{\"{O}stling Linear Regr.} & \textbf{\"{O}stling Logistic Regr.} & \textbf{Malaviya Linear Regr.} & \textbf{Malaviya Logistic Regr.} \\
    \midrule
        \textbf{Noun-adjective} & 0.146 & 0.261 & 0.141 & 0.378 \\ 
        \textbf{Noun-numeral} & 0.140 & 0.132 & 0.129 & 0.399 \\
        \textbf{Subject-verb} & 0.0781 & 0.306 & 0.101 & 0.156 \\ 
        \textbf{Object-verb} & 0.169 & 0.237 & 0.0757 & 0.122 \\ 
        \textbf{Object-subject} & 0.0127 & – & 0.0349 & 0.00940 \\ 
    \bottomrule
    \end{tabular}}
    \caption{Mean squared error scores for linear regression and logistic regression models for each feature, using language vectors from \citet{ostling-tiedemann-2017-continuous} and \citet{malaviya-etal-2017-learning}. Better scores are closer to 0.}
    \label{tab:mseScores}
\end{table*}

\begin{table*}[!ht]
    \centering
    \resizebox{0.94\textwidth}{!}{
    \begin{tabular}{l|cc|cc}
    \toprule
        & \textbf{\"{O}stling Linear Regr.} & \textbf{\"{O}stling Logistic Reg.} & \textbf{Malaviya Linear Regr.} & \textbf{Malaviya Logistic Regr.} \\
    \midrule
        \textbf{Noun-adjective} & -0.0423 & -1.41 & 0.0810 & -0.780 \\ 
        \textbf{Noun-numeral} & 0.246 & -3.15 & -14.0 & -2.45 \\ 
        \textbf{Subject-verb} & -0.233 & -1.21 & -0.627 & -0.776 \\ 
        \textbf{Object-verb} & -0.137 & -3.12 & 0.00891 & -0.486 \\ 
        \textbf{Object-subject} & -0.299 & – & -0.277 & -1.84 \\ 
    \bottomrule
    \end{tabular}}
    \caption{\(r^{2}\) scores for linear regression and logistic regression models for each feature, using language vectors from \citet{ostling-tiedemann-2017-continuous} and \citet{malaviya-etal-2017-learning}. Better scores are closer to 1.}
    \label{tab:r2scores}
\end{table*}

\section{Proposed Task and Model Comparison}

Because categorical typological datasets are a core part of many existing  typology-related NLP tasks, these tasks also suffer from many of the problems that the underlying datasets do. 
Examples of these tasks include typological feature prediction \citep{malaviya-etal-2017-learning,bjerva-etal-2020-sigtyp,10.1162/coli_a_00498}, low-resource language vocabulary prediction \citep{rani-etal-2023-findings}, and language identification from speech \citep{salesky-etal-2021-sigtyp}. 
It is for this reason that we introduce, along with the seed dataset, a new task predicting these novel continuous typological features.  
Unlike previous typological prediction tasks, the one we present here is regression-based.

\subsection{Methodological Comparison}
Most typological feature prediction (TFP) approaches use logistic regression  (e.g. \citealp{malaviya-etal-2017-learning,bjerva-augenstein-2018-phonology,bjerva_augenstein:18:iwclul,ostling-kurfali-2023}), as they are modelling categorical outcome variables.
However, we argue that linear regression is a more suitable method for TFP, since a more appropriate representation of typology is continuous \citep{levshina_2023}.
To quantify the differences between these approaches, we compare prediction results based on pretrained language vectors from \citet{ostling-tiedemann-2017-continuous} and \citet{malaviya-etal-2017-learning}.

As a baseline, we train logistic regression models on a discretized version of the word order features from our dataset. We have rounded each proportion to 0 or 1 (with all numbers 0.5 and above going to 1), to simulate a still-categorical version of the data, while ensuring comparability with the linear regression data. In this case, we use the following:

\[ \boldsymbol{Y} = \frac{1}{1+e^{(-\boldsymbol{\beta}\boldsymbol{X} - \boldsymbol{\beta}_{0})}} \]

where $ \boldsymbol{X} $ is a matrix made up of pretrained language vectors, $ \boldsymbol{Y} $ is a vector made up of the input language vectors' corresponding typological feature values, and $\boldsymbol{\beta}$ and $\boldsymbol{\beta}_{0}$ are the learned parameters. We employ the Scikit-learn \citep{scikit-learn,sklearn_api} implementation, which aims to find the optimal values of $\boldsymbol{\beta}$ and $\boldsymbol{\beta}_{0}$ by minimizing the log likelihood of the data. %We also used L2 regularization for these logistic regression models.

As an alternative approach, we train linear regression models on the language representations and use our gradient word order typology labels. For the modelling, we use:

\[ \boldsymbol{Y} = \boldsymbol{X}\boldsymbol{\beta} + \boldsymbol{\varepsilon} \]

where $ \boldsymbol{X} $ is again a matrix made up of pretrained language vectors, $ \boldsymbol{Y} $ is again a vector made up of the input language vectors' corresponding typological feature values, $ \boldsymbol{\beta} $ is the vector of learned regression coefficients, and $ \varepsilon $ is the bias vector. We use the Scikit-learn \citep{scikit-learn,sklearn_api} implementation of linear regression to train the model, which does so by minimizing the residual sum of squares between the real feature values and the predicted feature values.

For all models, both linear and logistic, we trained on a subset of the available languages, and display results, measured both in mean squared error and $ r^{2} $ score, calculated on a held-out test set. Because we employed pretrained language vectors as part of the training process, we were only able to train and evaluate each feature model on the set of languages that had both a pretrained language vector, and a value in our dataset for that feature. Unfortunately, this meant that our training set for each model had only around 40 datapoints, while our held-out evaluation set had only around 10 (with some slight variation depending on the feature and the language vector source). In cases where these languages had multiple available treebanks, we randomly selected one treebank to use, to avoid training on the same input vectors with potentially different expected output feature values. We selected one treebank randomly instead of combining them into one set per language so as to not arbitrarily combine data from potentially vastly different domains.
Detailed results are displayed in Tables \ref{tab:mseScores} and \ref{tab:r2scores}.

\subsection{Results and Discussion}
Given that the data at hand is continuous, and that linear regression models predict categorical values while logistic regression models predict binary values, we expected the linear regression models to outperform the logistic regression models on this task.
Indeed, the linear regression models perform better on average than the logistic regression models, when evaluated using mean squared error and \(r^{2}\) score. While not always the case, this is most often true as well on the individual feature level. 
While improvements to the modelling can be implemented, these baselines serve as an initial exploration of how to approach the novel task of regression-based typology prediction.

An important note from our statistical results is that the differences we observe between the data driven distributions and typological databases (Fig.~\ref{fig:featureDensity}) clearly show the limitations of established databases in terms of language descriptiveness on a fine-grained scale.
This discrepancy may to some extent explain the difficulty observed in empirical NLP experiments, when trying to integrate coarse-level WALS features in various NLP pipelines \citep{ponti2019modeling}.
The introduction of this regression-based typology prediction task may prove useful for incorporation of typological features in NLP modelling - for instance by incorporation as an auxiliary task.

While data-driven typology enables more fine-grained language description, it should be noted that the source of a treebank can have a considerable effect on the estimate \citep{levshina_2023}. \citet{baylor_underReview} show that linguistic variation, for instance stemming from domain, can affect word order values. 
Therefore, direct comparison between languages should ideally be based on parallel data.

\section{Conclusion}
\label{sec:conclusion}
Information from the field of linguistic typology has the potential to benefit the field of NLP. Unfortunately, the data from existing typological databases has been unreliable, largely due to their reliance on categorical features and those features' inability to represent the variability found in natural language. In this paper, we attempt to address this problem by introducing a new continuous-valued seed dataset, and argue that it is indeed better able to reflect the nuance of natural language when it comes to word order. In addition, we provide our dataset creation methodology that can be easily adapted in the future to generate data for a wider array of languages and features. Finally, we present a novel regression task based on predicting the feature values of this new dataset.

\section*{Limitations}
\label{sec:limitations}
The main limitation of our paper stems from the small size of our dataset, both in terms of number of features, and in terms of languages covered. As is always possible, our subset of features and languages could be misrepresentative of the larger existing features and languages, thus keeping our analyses from generalizing. The small size of our dataset only makes this more probable. 

A secondary limitation of this work primarily applies to our dataset creation method. As it currently stands, the method only works with annotated linguistic data, vastly cutting down on the amount of available useful language data.

\section*{Ethics Statement}
As this paper relies on existing linguistic data sources from which to generate datasets, no human data was collected. 

We do not foresee this work directly creating any substantial ethical issues, but we do note that language communities can be significantly impacted, both positively and negatively, by language technologies. Given that this research has the potential to aid in the further development of language technologies, we want to highlight the importance of community-led development, including ceasing development of technologies for certain languages based on community request.

\section*{Acknowledgements}
This work was supported by the Carlsberg Foundation under the \textit{Semper Ardens: Accelerate} programme (CF21-0454). 
EB was further supported by the McGill University Graduate Mobility Award to travel to AAU to carry out this work.

\bibliography{anthology,custom,generic_custom}
\bibliographystyle{acl_natbib}

\appendix
\newpage
\section{Tags for algorithm beyond Adjective-Noun order}
\label{sec:appendix}

\begin{table}[h]
\begin{tabular}{l|ll}
\toprule
\textbf{POS} & \textbf{UD \texttt{upos} value} & \textbf{UD \texttt{deprels} value} \\
\midrule
Noun                    & \texttt{NOUN}                     & –                           \\
Adjective               & \texttt{ADJ}                      & \texttt{amod}                        \\
Numeral                 & \texttt{NUM}                      & \texttt{nummod}                      \\
Subject                 & –                        & \texttt{nsubj}                       \\
Object                  & –                        & \texttt{obj}                         \\
Verb                    & \texttt{VERB}                     & –                  \\
\bottomrule
\end{tabular}
\caption{Tags used to extract the necessary parts of speech from the Universal Dependencies treebank \citep{nivre-etal-2020-universal}. Dashes indicate that that value did not need to be specified.}
\label{tab:ud-tags}
\end{table}

%\section{Geographical distribution of Adjective-noun gradient}
\label{sec:appendix-map}

%\begin{figure}[h]
%   % \centering
%    \includegraphics[width=0.5\textwidth]{Images/adj_noun_map_gradient.pdf}
 %   \caption{Distribution of gradient values for adjective-noun order in Western-European geographical area.}
%    \label{fig:map}
%\end{figure}

\end{document}